\documentclass{article}
\usepackage{spconf,amsmath,graphicx,hyperref}
\usepackage{array}
\usepackage{booktabs}
\usepackage{tabularx} 
\usepackage{graphicx} 
\usepackage{subcaption}
\usepackage{pifont}
\usepackage{amsthm} 
\usepackage{multirow}
\usepackage{amssymb} 
\usepackage[skip=2pt]{caption}

\title{DiaCDM: Cognitive Diagnosis in Teacher-Student Dialogues using the Initiation-Response-Evaluation Framework}
\name{\begin{tabular}{c}
Rui Jia$^{1,2}$\sthanks{Equal contribution},
Yuang Wei$^{1,2*}$, 
Ruijia Li$^{1,2}$, 
Yuan-Hao Jiang$^{1,2}$,
Xinyu Xie$^{1,2}$,\\
Yaomin Shen$^{3}$,
Min Zhang$^{1,2}$\sthanks{Project leader}\sthanks{Corresponding authors. Min Zhang (mzhang@cs.ecnu.edu.cn) and Bo Jiang (bjiang@deit.ecnu.edu.cn) }, 
Bo Jiang$^{1,2\ddagger}$
\end{tabular}}

\address{
$^1$Lab of Artificial Intelligence for Education, East China Normal University, China\\
$^2$Shanghai Institute of Artificial Intelligence for Education, East China Normal University, China\\
$^3$XR System Application Research Center, Nanchang Research Institute, Zhejiang University, China
}

\begin{document}
\ninept
\maketitle
%



\begin{abstract}
While cognitive diagnosis (CD) effectively assesses students' knowledge mastery from structured test data, applying it to real-world teacher-student dialogues presents two fundamental challenges. Traditional CD models lack a suitable framework for handling \textit{dynamic, unstructured dialogues}, and it's difficult to accurately \textit{extract diagnostic semantics from lengthy dialogues}. To overcome these hurdles, we propose DiaCDM, an innovative model. We've adapted the initiation-response-evaluation (IRE) framework from educational theory to design a diagnostic framework tailored for dialogue. We also developed a unique \textbf{graph-based encoding} method that integrates teacher questions with relevant knowledge components to capture key information more precisely. To our knowledge, this is the first exploration of cognitive diagnosis in a dialogue setting. Experiments on three real-world dialogue datasets confirm that DiaCDM not only significantly improves diagnostic accuracy but also enhances the results' interpretability, providing teachers with a powerful tool for assessing students' cognitive states. The code is available at \url{https://github.com/Mind-Lab-ECNU/DiaCDM/tree/main}.

\end{abstract}
\begin{keywords}
Cognitive diagnosis, Teacher-student dialogue CD, IRE dialogue framework, Graph-based encoding  
\end{keywords}
\section{Introduction}
\label{sec:intro}

\begin{figure}[tb]
    \centering
    \noindent
    \includegraphics[width=\columnwidth]{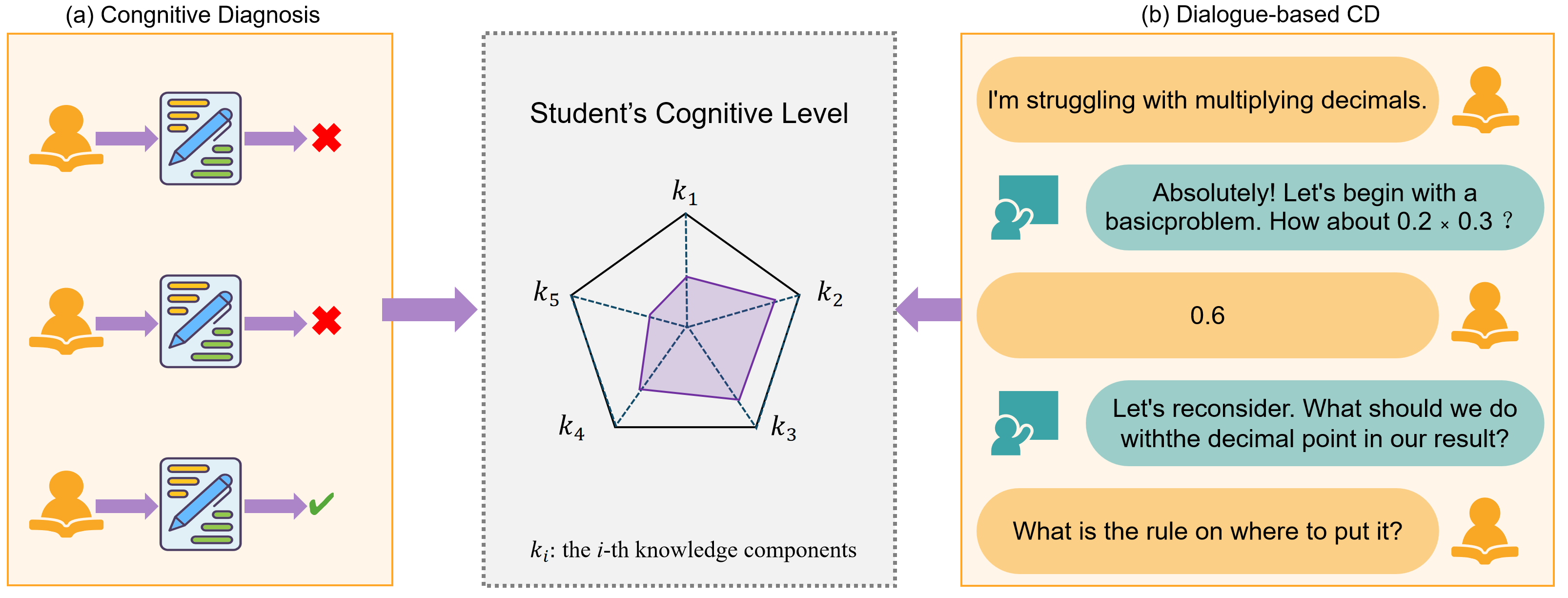}
    \caption{(a) \textbf{Cognitive diagnosis}: Diagnose students’ cognitive level of knowledge components through structured test items. (b) \textbf{Dialogue-based CD}: Dynamically identify students’ cognition of knowledge components by analyzing teacher-student dialogues.}
    \label{fig:toy}
\end{figure}

Cognitive Diagnosis (CD) aims to infer students’ mastery of knowledge components based on their historical learning performance and records. Most existing cognitive diagnosis models (CDM) rely on item response theory(IRT) method, focusing on static and single-round test settings (as shown in Fig.~\ref{fig:toy}(a))~\cite{survey,DiaMeaning,ICD}. However, in real educational environments, dynamic teacher-student dialogue scenarios play an equally critical for identifying students’ cognitive states and advancing personalized instruction (as shown in Fig.~\ref{fig:toy}(b))~\cite{budzianowski2019hello,xu2023improving}.

Recently, some researchers have attempted to leverage large language models (LLMs) to generate advanced semantic encodings of dynamic teacher-student dialogue data, and feed these encodings into cognitive diagnosis models (CDMs) to address this challenge~\cite{jiangExplainableLearningOutcomes2025,jiangMASKCLKnowledgeComponent2025b,ncd}. However, we observe that in real classroom settings, teacher utterances are \textbf{typically long and information-dense}, whereas student \textbf{responses are short and highly context-dependent}. Applying the same encoding to both teachers and students may overlook critical information in student responses and underestimate the role of teacher scaffolding and feedback, both of which are essential for accurate cognitive diagnosis and personalized instruction.

Inspired by Abstract Meaning Representation (AMR)~\cite{AMR_survey}, which encodes text into semantic graphs to highlight core semantics and suppress irrelevant tokens, AMR has demonstrated remarkable effectiveness in tasks such as reading comprehension~\cite{machine1,machine2}, summarization~\cite{summ1,summ2} and machine translation~\cite{trans1,trans2,trans3}. Building on this, we propose a dialogue-based cognitive diagnosis model, DiaCDM. Specifically, the model first decomposes teacher–student dialogues into three components—initiation, response, and evaluation—under the Initiation–Response–Evaluation (IRE) framework. Then, teacher questions are encoded into semantic graphs via AMR, from which an attention-enhanced Graph Convolutional Network (GCN) extracts core semantics and their associations with knowledge components. Finally, DiaCDM jointly models students’ cognitive states from three perspectives: question matching, response-level cognition, and evaluation-level cognition. Experiments on three dialogue datasets show that DiaCDM significantly outperforms traditional test-based cognitive diagnosis models. Furthermore, our interpretability analysis reveals how different learning factors shape students’ cognitive processes and states, not only maintaining strong predictive performance but also providing actionable insights for instructional practice. The contributions of our paper are as follows:
\begin{itemize}
    \item We systematically extend CD from tests to teacher-student dialogues, identifying two key challenges: the lack of dialogue frameworks and difficulty in extracting diagnostic semantics.
    \item We propose DiaCDM, which adapts the IRE and introduces AMR-based graph representations with attention to effectively model dialogues and enhance semantic representations.
    \item Extensive experiments on three real-world dialogue datasets show that our DiaCDM significantly outperforms existing methods in both diagnostic accuracy and interpretability.
\end{itemize}

\section{related work}
CDMs aim to infer students' mastery of knowledge components, and existing approaches are mainly divided into traditional psychometric methods and machine learning methods. The former, represented by IRT~\cite{irt} and its multidimensional extensions, characterize students' cognitive states through latent variables; the latter leverage deep learning to capture complex student–item interactions and further incorporate multimodal information, such as Neural Cognitive Diagnosis Model (NCDM)~\cite{ncd} and Knowledge-Association Neural Cognitive Diagnosis Model (KaNCD)~\cite{kancd}, as well as DIRT~\cite{dirt} and NCD+~\cite{ncd}. However, these methods are mostly limited to static, test-based settings and are difficult to adapt to the dynamic interactions of real teaching. Teacher-student dialogue, as a typical interactive scenario, usually follows the IRE (Initiation–Response–Evaluation) structure~\cite{mehan1980competent}. Teachers' questions and feedback, as well as students' responses, contain rich and diverse diagnostic signals, and studies have shown that high-quality dialogues can effectively reveal and influence students' cognitive processes. Nevertheless, these potential diagnostic signals have not been fully utilized in existing dialogue-based tutoring systems, which often focus more on dialogue generation and interaction strategies, lacking systematic and fine-grained student modeling. Therefore, extending CDMs to dialogue scenarios and achieving cognitive diagnosis from dynamic interactions remains an important yet challenging research direction. Although recent studies have attempted to incorporate knowledge tracing tasks into teacher–student dialogue settings~\cite{scarlatos2025exploring}, achieving accurate cognitive diagnosis in open-ended interactions still poses a significant challenge for CDM models.

\section{Preliminary}
\textbf{Problem Definition.}  
Given the dialogue logs $\mathcal{L}$ of students and the associated knowledge components, the goal of cognitive diagnosis is to infer students' mastery levels by modeling their performance during dialogues. Let $\mathcal{S} = \{s_1, \dots, s_{|\mathcal{S}|}\}$ and $\mathcal{K} = \{k_1, \dots, k_{|\mathcal{K}|}\}$ denote the sets of students and knowledge components, respectively, where $|\cdot|$ is the set size. Each student $s_i$ engages in $M_i$ dialogue rounds with a teacher, and the dialogue logs are represented:
\begin{equation}
\mathcal{L} = \{ x_{ij} \mid i = 1, \dots, |\mathcal{S}|, \; j = 1, \dots, M_i \},
\end{equation}
where each \(x_{ij} = (q_{ij}, a_{ij}, e_{ij}, r_{ij})\) clearly consists of the teacher's question \(q_{ij}\), the student's answer \(a_{ij}\), the teacher's evaluation \(e_{ij}\), and the binary correctness \(r_{ij} \in \{0,1\}\). For each given question $q_{ij}$, the related concepts then form the set $\mathcal{K}_{ij} \subseteq \mathcal{K}$:

\begin{equation}
\mathcal{K}_{ij} = \Big\{ k_x \;\Big|\; k_x \in \mathcal{K}, \; f(q_{ij}, k_x) = 1 \Big\},
\end{equation}
where $f(q_{ij}, k_x)$ is an indicator function with $f(q_{ij}, k_x)=1$ if $q_{ij}$ involves $k_x$, and $f(q_{ij}, k_x)=0$ otherwise. We make two common assumptions in cognitive diagnosis: (\textit{i}) a student's cognitive state remains stable during a dialogue session, as ability does not change significantly over short periods, and (\textit{ii}) the probability of answering correctly increases monotonically with mastery of the relevant concepts, consistent with intuition and empirical evidence.

\noindent \textbf{Data Pre-processing.}  
Following the well-established IRE framework, teacher-student dialogues are systematically reorganized into the canonical IRE structure for accurate subsequent modeling.
\section{Method}
\begin{figure*}[tb]
  \centering
  \includegraphics[width=\textwidth]{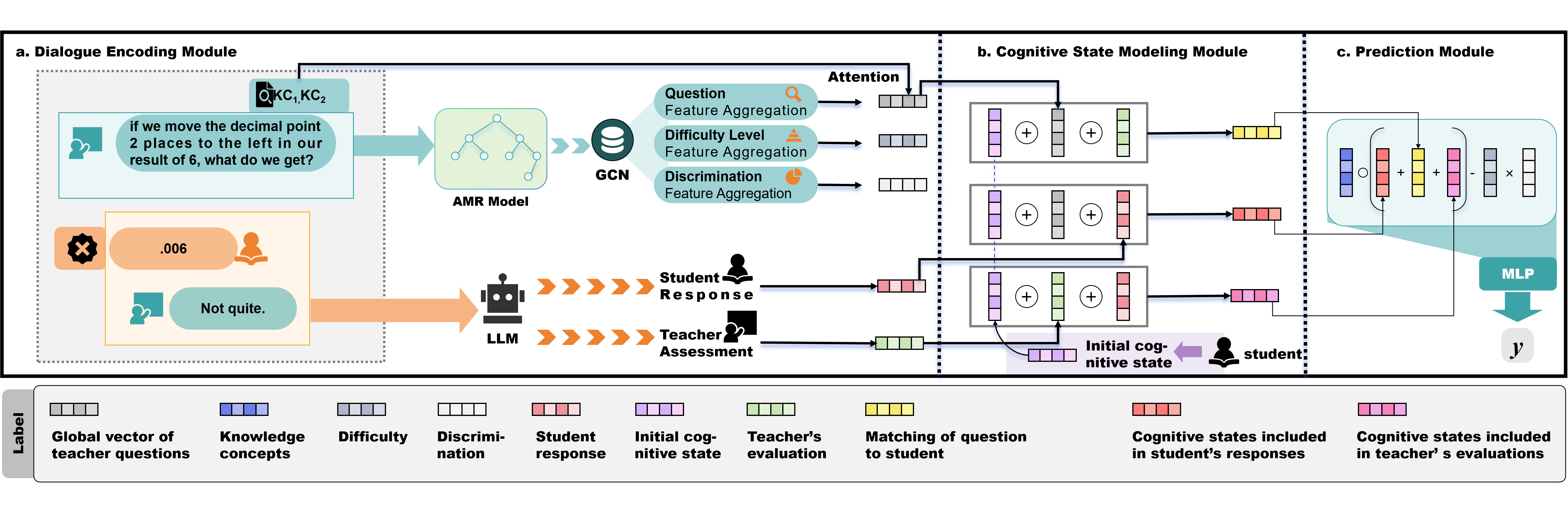}
  \caption{Overview of DiaCDM: (a) Encode the teacher's questions, student responses, and teacher evaluations separately. (b) Capture the matching of the teacher's questions, the cognitive state in student responses, and the cognitive state in teacher evaluations from the encoded vectors. (c) Predict the student's future performance by integrating the information from these three dimensions with educational theory.}
  \label{fig:framework}
\end{figure*}

In this section, we present the proposed dialogue-based cognitive diagnosis model, \textbf{DiaCDM} (Fig.~\ref{fig:framework}), specifically designed to infer students' knowledge mastery from teacher-student dialogues. DiaCDM consists of three core modules: the \textbf{Dialogue Encoding Module} extracts semantic features from teacher questions, student responses, and teacher evaluations, while integrating relevant knowledge components; the \textbf{Cognitive State Modeling Module} generates student cognitive representations, effectively capturing information reflected in both teacher questions and student answers; and the \textbf{Prediction Module} leverages the Q-matrix and question parameters to predict student performance, simultaneously refining knowledge representations to improve diagnostic accuracy and interpretability.

\subsection{Dialogue Encoding Module}
Teacher-student dialogues contain rich semantic information. To effectively extract the core semantics of a teacher's question $q$, we first generate its AMR graph using a pre-trained AMR model:

\begin{equation}
G_q = \text{AMR}(q) = (\mathrm{Nod}_q, \mathrm{Edg}_q),
\end{equation}
where $\mathrm{Nod}_q = \{ n_1, \dots, n_{|\mathrm{Nod}_q|} \}$ and $\mathrm{Edg}_q$ denote the nodes and edges capturing semantic units and their relationships. Each node is encoded into a semantic vector using a pre-trained LLM:
\begin{equation}
\mathbf{h}_i^{(0)} = \mathrm{LLM}(n_i), \quad i = 1, \dots, |\mathrm{Nod}_q|.
\end{equation}

Three independent GCNs then propagate and aggregate node features, each explicitly focusing on a distinct aspect of the question: global semantics, inherent difficulty, and discrimination:
\begin{equation}
\mathbf{h}_{g_i}^{(l)} = \sigma(\mathbf{\tilde{D}}^{-\frac12}\mathbf{\tilde{A}}\mathbf{\tilde{D}}^{-\frac12}\mathbf{h}_{g_i}^{(l-1)} \mathbf{W}_g^{(l-1)}),
\end{equation}
\begin{equation}
\mathbf{h}_{f_i}^{(l)} = \sigma(\mathbf{\tilde{D}}^{-\frac12}\mathbf{\tilde{A}}\mathbf{\tilde{D}}^{-\frac12}\mathbf{h}_{f_i}^{(l-1)} \mathbf{W}_f^{(l-1)}),
\end{equation}
\begin{equation}
\mathbf{h}_{d_i}^{(l)} = \sigma(\mathbf{\tilde{D}}^{-\frac12}\mathbf{\tilde{A}}\mathbf{\tilde{D}}^{-\frac12}\mathbf{h}_{d_i}^{(l-1)} \mathbf{W}_d^{(l-1)}),
\end{equation}
where $\mathbf{\tilde{A}} = \mathbf{A} + \mathbf{I}$ is the adjacency matrix with self-loops, $\mathbf{\tilde{D}}$ is its corresponding degree matrix, $\mathbf{W}_g^{(l)}, \mathbf{W}_f^{(l)}, \mathbf{W}_d^{(l)}$ are trainable weight parameters, and $\sigma$ is a standard nonlinear activation function.

After GCN layers, the learned node features are iteratively propagated, aggregated, and linearly projected to obtain the final comprehensive question vectors for subsequent cognitive state modeling:

\begin{equation}
\mathbf{\tilde{h}}_g = W_1 \mathbf{h}_g + b_1, \quad
\mathbf{\tilde{h}}_f = W_2 \mathbf{h}_f + b_2, \quad
\mathbf{\tilde{h}}_d = W_3 \mathbf{h}_d + b_3,
\end{equation}
where $\mathbf{\tilde{h}}_g$, $\mathbf{\tilde{h}}_f$, and $\mathbf{\tilde{h}}_d$ respectively represent the final global semantics, inherent difficulty, and discrimination aspects, and $W_1, W_2, W_3$ along with $b_1, b_2, b_3$ are learnable projection weight parameters.

Teacher questions often involve multiple knowledge components with varying importance. Each concept $k_i$ is first encoded:
\begin{equation}
\mathbf{h}_k^{(i)} = \mathrm{LLaMA}(k_i), \quad i = 1, \dots, |K_q|,
\end{equation}
where $|K_q|$ is the total number of concepts, and $\mathbf{h}_k^{(i)} \in \mathbb{R}^{dim_g}$. A standard attention mechanism then dynamically and efficiently fuses the question vector $\mathbf{\tilde{h}}_g$ with the complete set of concept vectors $\{\mathbf{h}_k^{(i)}\}$ to produce a knowledge-weighted global semantic vector:

\begin{equation}
\mathbf{\tilde{h}}_{gk} = \mathrm{Attention}(\mathbf{\tilde{h}}_g, \mathbf{h}_k).
\end{equation}

Student responses $a$ and teacher evaluations $e$ are generally short, concise, and contextually rich, and are directly encoded using a pre-trained language model for efficient semantic representation:

\begin{equation}
\mathbf{h}_a = \mathrm{LLaMA}(a), \quad \mathbf{h}_e = \mathrm{LLaMA}(e),
\end{equation}
where $\mathbf{h}_a, \mathbf{h}_e \in \mathbb{R}^{dim_g}$ are their semantic vectors. These vectors are then seamlessly integrated with the question representations to infer students' cognitive states and support downstream prediction tasks.

\subsection{Cognitive State Modeling Module}

In a dialogue, teachers continuously assess students' cognitive states by asking questions, to which students respond and subsequently receive teacher feedback. We model students' cognitive states from three complementary perspectives: (1) matching between teacher questions and students' current cognitive states, $C_q$, (2) cognitive states inferred by the teacher from student responses, $C_t$, and (3) cognitive states accurately reflected in student responses, $C_s$.

To capture individual differences, each student is assigned a unique initial cognitive state vector $\mathbf{h}_s \in \mathbb{R}^{1 \times |K|}$, initialized as:
\begin{equation}\label{eq:initial_state}
\mathbf{h}_s = \mathbf{s} \mathbf{A},
\end{equation}
where $\mathbf{s} \in \mathbb{R}^1$ is the student identifier, $\mathbf{A} \in \mathbb{R}^{1 \times |K|}$ is a learnable matrix, and $|K|$ denotes the total number of knowledge components.

The alignment between teacher questions and student cognitive states, $C_q \in \mathbb{R}^{1 \times |K|}$, is explicitly determined by the initial state $\mathbf{h}_s$, the question's extracted global semantic vector $\mathbf{\tilde{h}}_{gk}$, and teacher feedback $\mathbf{h}_e$, and is computed via a multi-layer perceptron (MLP):

\begin{equation}\label{eq:alignment_question}
C_q = \mathbf{W}_q \cdot \text{Concat}(\mathbf{h}_s, \mathbf{\tilde{h}}_{gk}, \mathbf{h}_e) + \mathbf{b}_q,
\end{equation}

The cognitive state inferred from student responses, $C_t \in \mathbb{R}^{1 \times |K|}$, directly depends on the initial student state $\mathbf{h}_s$, the corresponding student response vector $\mathbf{h}_a$, and teacher feedback $\mathbf{h}_e$:

\begin{equation}\label{eq:inferred_state}
C_t = \mathbf{W}_t \cdot \text{Concat}(\mathbf{h}_s, \mathbf{h}_a, \mathbf{h}_e) + \mathbf{b}_t,
\end{equation}

Finally, the cognitive state in student responses, $C_s \in \mathbb{R}^{1 \times |K|}$, is determined by the student's initial state $\mathbf{h}_s$, the corresponding question vector $\mathbf{\tilde{h}}_{gk}$, and the observed student response $\mathbf{h}_a$:

\begin{equation}\label{eq:reflected_state}
C_s = \mathbf{W}_s \cdot \text{Concat}(\mathbf{h}_s, \mathbf{\tilde{h}}_{gk}, \mathbf{h}_a) + \mathbf{b}_s,
\end{equation}
where $\mathbf{W}_q, \mathbf{W}_t, \mathbf{W}_s \in \mathbb{R}^{3 \cdot dim_g \times |K|}$ and $\mathbf{b}_q, \mathbf{b}_t, \mathbf{b}_s$ are learnable trainable parameters for weight and bias, respectively, in the model.

\subsection{Prediction Module}
To predict students' future performance, we leverage DINA~\cite{dina} and IRT~\cite{irt} theories for enhanced expressiveness and interpretability. DINA's Q-matrix encodes the mapping between questions and knowledge components, while IRT models account for both student ability and question properties. Accordingly, the set of knowledge components in each teacher question is converted into a one-hot vector \(K_q \in \{0,1\}^{|K|}\), where \(|K|\) is the total number of concepts.

The three cognitive vectors obtained previously (\(C_q, C_t, C_s\)) are carefully and systematically combined into a single comprehensive cognitive representation \(h_c\) via a weighted sum, effectively capturing students' overall mastery of the relevant knowledge components:

\begin{equation}
h_c = \lambda_1 C_q + \lambda_2 C_t + (1-\lambda_1-\lambda_2) C_s,
\end{equation}
where \(\lambda_1\) and \(\lambda_2\) are learnable weights, which are optimized during model training to balance the contributions of each cognitive vector.

Prediction is then computed element-wise using the question vector and the corresponding transformed feature representations:
\begin{equation}
x = (h_c - \tilde{h}_f) \circ \tilde{h}_d, \quad \hat{y} = \sigma(W_2 \cdot (W_1 \cdot x + b_1) + b_2),
\end{equation}
where \(\circ\) denotes element-wise multiplication, and \(\sigma\) is the Sigmoid function. The model is trained via cross-entropy loss:
\begin{equation}
\mathcal{L} = - \sum \left[y \log \hat{y} + (1-y) \log (1-\hat{y})\right].
\end{equation}

After training, \(h_c\) serves as the final diagnostic vector representing each student's mastery over the relevant knowledge components.

\section{Experiments}
We conduct experiments on three real-world datasets to evaluate DiaCDM. Our experiments aim to answer the following questions: \textbf{RQ1}: Does DiaCDM perform better than existing CDMs in dialogue settings? \textbf{RQ2}: Do the different components of DiaCDM contribute to its performance? \textbf{RQ3}: Does DiaCDM exhibit interpretability?

\noindent \textbf{Datasets.}
Following~\cite {scarlatos2025exploring}, we conducted experiments using three datasets: CoMTA, MathDial, and PMTD. These datasets cover human-AI and teacher-student dialogue interactions, demonstrating the applicability of our approach across diverse dialogue settings.

\noindent \textbf{Evaluation metrics.}
Evaluating CDMs is challenging since students' actual mastery of knowledge points cannot be directly observed. A common approach is to predict their future performance as a reliable proxy for knowledge level. Models are trained to predict whether a student can correctly solve a problem, with metrics such as Area Under the Curve (AUC) and Accuracy (ACC) assessing prediction accuracy. This prediction-based evaluation provides an indirect yet effective assessment of students' cognitive states.

\noindent \textbf{Implementation details.}
During training, all model parameters are Xavier-initialized and optimized with Adam at a learning rate of 0.002 and batch size 64. We use LLAMA-7B as the LLM and AMR3-structbart-L~\footnote{\url{https://github.com/IBM/transition-amr-parser}} as the pre-trained AMR model. ChatGPT-4o accurately transforms dialogues into IRE-structured sequences. Following \cite{li2024towards, gao2024zero}, datasets are randomly split 8:1:1 for training, validation, and testing. All experiments are repeated five times with different random seeds, and the reported averages are shown.

\noindent \textbf{Baseline.}  
We first compared DiaCDM with classical CDMs (DINA, IRT, MIRT, MCD, NCD, KaNCD, SymbolicCDM) on three real-world dialogue datasets. Since these models cannot directly process problem text, we designed two simple baselines:  
1) \textbf{base(B)}: Classical CDMs using only item attributes;  
2) \textbf{text(T)}: CDMs enhanced with problem text, encoded in the same way as in DiaCDM.

\begin{table}[tb]
    \centering
    \setlength{\tabcolsep}{2pt} 
    \caption{Results on three dialogue datasets.}
    \label{tab:results}
    \begin{tabular}{lccccccccc}
    \toprule
         \multicolumn{2}{c}{\multirow{2}{*}{Methods}} & \multicolumn{2}{c}{CoMTA} & \multicolumn{2}{c}{MathDial} & \multicolumn{2}{c}{PMTD} \\
         \cmidrule(lr){3-4} \cmidrule(lr){5-6} \cmidrule(lr){7-8}
         & & AUC & ACC & AUC & ACC & AUC & ACC \\
         \midrule
         \multirow{2}{*}{DINA~\cite{dina}} & B & 0.546 & 0.538 & 0.553 & 0.553 & 0.535 & 0.561 \\
         & T & 0.612 & 0.596 & 0.556 & 0.576 & 0.510 & 0.527 \\
         \midrule
         \multirow{2}{*}{IRT~\cite{irt}} & B & 0.569 & 0.560 & 0.573 & 0.525 & 0.510 & 0.547 \\
         & T & 0.588 & 0.553 & 0.602 & 0.558 & 0.526 & 0.656 \\
         \midrule
         \multirow{2}{*}{MIRT~\cite{mirt}} & B & 0.560 & 0.547 & 0.582 & 0.561 & 0.524 & 0.561 \\
         & T & 0.582 & 0.564 & 0.526 & 0.539 & 0.558 & 0.531 \\
         \midrule
         \multirow{2}{*}{MCD~\cite{MFNCD}} & B & 0.636 & 0.585 & 0.562 & 0.543 & 0.529 & 0.577 \\
         & T & \underline{0.652} & 0.574 & 0.599 & 0.562 & 0.551 & 0.622 \\
         \midrule
         \multirow{2}{*}{NCD~\cite{ncd}} & B & 0.591 & 0.553 & 0.554 & 0.540 & 0.519 & \underline{0.656} \\
         & T & 0.637 & 0.553 & 0.590 & 0.565 & \underline{0.566} & \underline{0.656} \\
         \midrule
         \multirow{2}{*}{KaNCD~\cite{kancd}} & B & \underline{0.692} & \underline{0.655} & \underline{0.610} & 0.571 & \underline{0.569} & 0.595 \\
         & T & 0.656 & 0.553 & 0.599 & \underline{0.579} & 0.545 & 0.593 \\
         \midrule
         \multirow{2}{*}{SymCDM~\cite{symbolic}} & B & 0.677 & 0.627 & 0.531 & 0.544 & 0.525 & 0.525 \\
         & T & 0.684 & \underline{0.693} & 0.547 & 0.513 & \textbf{0.667} & 0.530 \\
         \midrule
         DiaCDM-Qwen & - & \underline{0.750} & \underline{0.681} & 0.722 & \underline{0.675} & 0.614 & 0.646 \\
         DiaCDM-DS & - & 0.662 & 0.528 & \underline{0.725} & 0.630 & 0.633 & \underline{0.660} \\
         DiaCDM & - & \textbf{0.789} & \textbf{0.728} & \textbf{0.845} & \textbf{0.774} & \textbf{0.704} & \textbf{0.704} \\
         \bottomrule
    \end{tabular}
\end{table}

\noindent \textbf{Comparative study.}  
In our comparative study, we first evaluated DiaCDM against our baselines. We then examined the impact of different LLM for text encoding, including LLaMA-7B (DiaCDM), Qwen-2.5-7B (DiaCDM-Qwen)~\cite{qwen}, and DeepSeek-Math-7B (DiaCDM-DS)~\cite{deepseek}. Table~\ref{tab:results} shows that in dialogue settings, these baselines, even with question text, perform far worse than DiaCDM. Among the LLM variants, LLaMA achieves the best results, reflecting its stronger ability to capture problem semantics and dialogue-specific information. Overall, while textual semantics improve performance, DiaCDM's main advantage comes from its multidimensional modeling of dialogue-specific cognitive signals, which is crucial for accurate and interpretable cognitive diagnosis.

\noindent \textbf{Ablation study.}
We conducted ablations on DiaCDM: \textbf{w/o-AMR} (DiaCDM without AMR, using only the LLM to encode dialogue), \textbf{w/o-KC} (ignoring knowledge component emphasis in teacher questions), \textbf{w/o-qM}, \textbf{w/o-ts}, and \textbf{w/o-se} (excluding question matching, student response states, and teacher evaluation states, respectively). Table~\ref{tab:combined_comparison} shows that removing AMR causes the largest drop in PMTD, while omitting question matching most affects CoMTA and MathDial, highlighting the importance of semantic extraction and alignment. Compared with text-augmented baselines (DeepIRT+, NCD+), DiaCDM's improvements mainly come from IRE theory and multidimensional cognitive modeling.

\begin{table}[tb]
    \centering
    \caption{Ablation study of DiaCDM}
    \label{tab:combined_comparison}
    \setlength{\tabcolsep}{4pt}
    \renewcommand{\arraystretch}{1.1}
    \begin{tabularx}{\columnwidth}{l *{6}{>{\centering\arraybackslash}X}}
        \toprule
        \multirow{2}{*}{Model} & \multicolumn{2}{c}{CoMTA} & \multicolumn{2}{c}{MathDial} & \multicolumn{2}{c}{PMTD} \\
        \cmidrule(lr){2-3} \cmidrule(lr){4-5} \cmidrule(lr){6-7}
        & AUC & ACC & AUC & ACC & AUC & ACC \\
        \midrule
        W/o AMR & \underline{0.7657} & \underline{0.6936} & \underline{0.7453} & \underline{0.6739} & 0.5926 & 0.6870 \\
        W/o KC  & 0.7564 & 0.5531 & 0.7408 & 0.3677 & \underline{0.6505} & \underline{0.7251} \\
        W/o qM  & 0.6708 & 0.6425 & 0.6677 & 0.6121 & 0.6445 & 0.7129 \\
        W/o ts  & 0.6980 & 0.6255 & 0.7102 & 0.6566 & 0.6097 & 0.5428 \\
        W/o se  & 0.7327 & 0.6000 & 0.7231 & 0.6546 & 0.5903 & 0.6108 \\
        DeepIRT+ & 0.6857 & 0.6489 & 0.6264 & 0.5789 & 0.5612 & 0.5510 \\
        NCD+      & 0.5748 & 0.5531 & 0.5192 & 0.4560 & 0.5893 & 0.6559 \\
        DiaCDM  & \textbf{0.7886} & \textbf{0.7276} & \textbf{0.7600} & \textbf{0.7000} & \textbf{0.7225} & \textbf{0.7278} \\
        \bottomrule
    \end{tabularx}
\end{table}

\begin{figure}[tb]  
  \centering
  \includegraphics[width=\columnwidth]{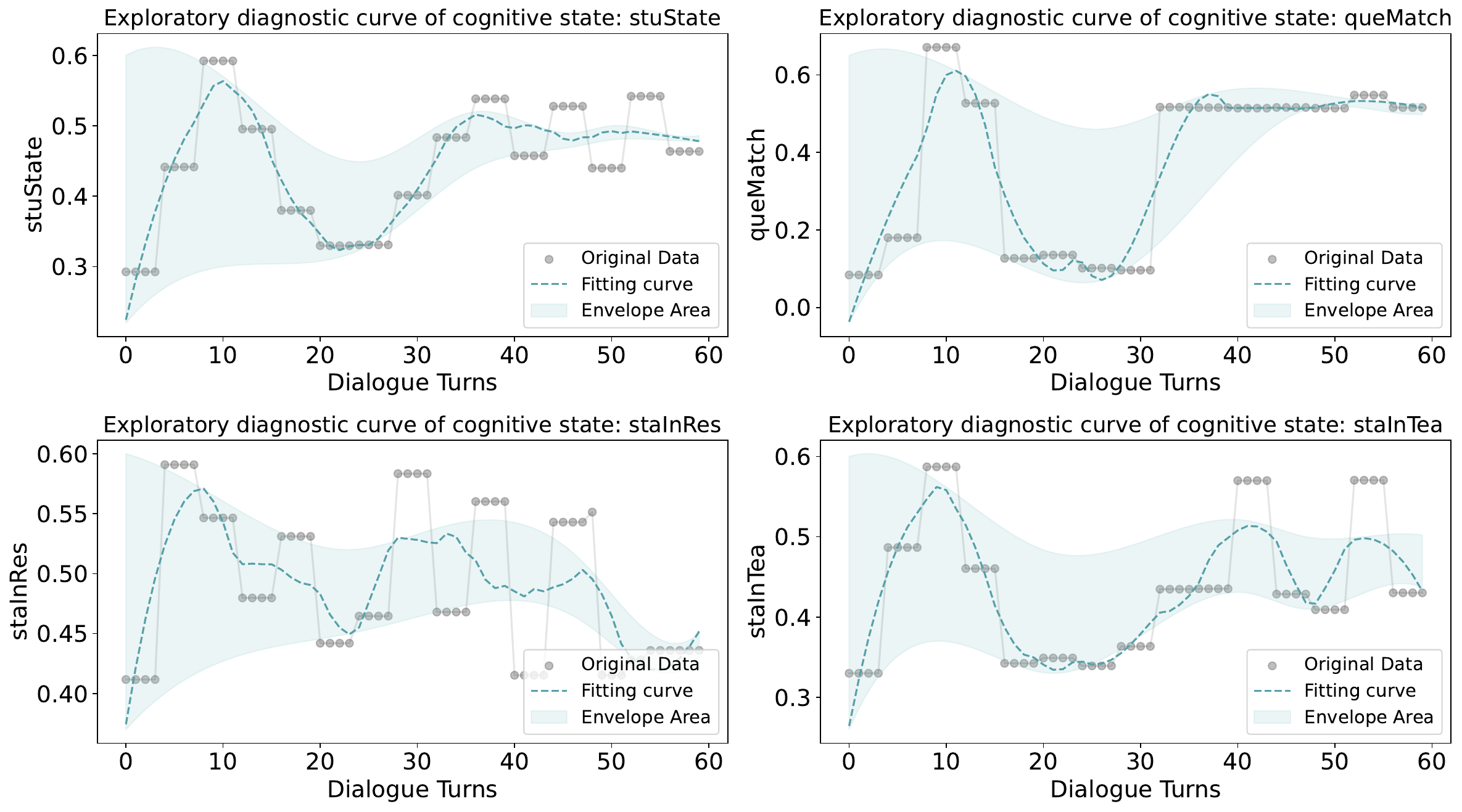}
  \caption{Visualization of Main and Sub-feature Dynamics in Cognitive State Identification.}
  \label{fig:ex1}
\end{figure}

\noindent \textbf{Interpretability analysis.}
Fig.~\ref{fig:ex1} shows the evolution of the main feature \textbf{stuState} and three sub-features \textbf{queMatch}, \textbf{staInRes}, and \textbf{staInTea} during progressive identification of a student's current cognitive state. The main feature \textbf{stuState} reflects the model's overall evaluation, initially fluctuating due to limited information and stabilizing as evidence gradually accumulates. The sub-feature \textbf{queMatch} measures alignment between student answers and questions, clarifying understanding. \textbf{staInRes} captures application of knowledge points in responses, indicating mastery relative to question demands. \textbf{staInTea} reflects cues from the teacher's language, helping the model verify or refine its assessment of the student's state.

\section{Conclusion}
DiaCDM is a dialogue-based cognitive diagnosis method using the IRE framework. It encodes teacher questions as AMR graphs, emphasizes key concepts via attention, and represents student responses and teacher evaluations with a LLM. Integrating IRE theory, DiaCDM infers students' cognitive states and predicts performance. Experiments on three datasets show superior accuracy and interpretability over test-based methods. Future work will refine behavior modeling and extend DiaCDM to practical dialogue-based tutoring.


\bibliographystyle{IEEEbib}
\bibliography{strings,refs}

\end{document}